\documentclass[letterpaper, 10 pt, conference]{ieeeconf}  

\IEEEoverridecommandlockouts                              

\usepackage{graphicx}
\usepackage{xcolor}
\definecolor{mydarkblue}{rgb}{0,0.08,0.45}
\usepackage[colorlinks = true,
            linkcolor = mydarkblue,
            urlcolor  = mydarkblue,
            citecolor = mydarkblue,
            anchorcolor = mydarkblue]{hyperref}
\usepackage{amssymb}
\usepackage{glossaries}

\newacronym{rl}{RL}{Reinforcement Learning}
\newacronym{il}{IL}{Imitation Learning}
\newacronym{urma}{URMA}{Unified Robot Morphology Architecture}
\newacronym{ppo}{PPO}{Proximal Policy Optimization}
\newacronym{gru}{GRU}{Gated Recurrent Unit}
\newacronym{mse}{MSE}{Mean Squared Error}
\newacronym{mlp}{MLP}{Multilayer Perceptron}

\title{\LARGE \bf
Active Embodiment Identification with Reinforcement Learning\\
for Legged Robots
}

\author{
  Nico Bohlinger$^{1}$,
  Jan Peters$^{1,2}$
\thanks{
This project was funded by National Science Centre, Poland, under the OPUS call in the Weave program UMO-2021/43/I/ST6/02711, and by the German Science Foundation (DFG) under grant number PE 2315/17-1.
\newline
$^{1}$Department of Computer Science, Technical University of Darmstadt, Germany.
$^{2}$Robotics Institute Germany (RIG); German Research Center for AI (DFKI); hessian.AI.
\newline Corresponding author: \tt\small nico.bohlinger@tu-darmstadt.de
}}

\begin{document}

\maketitle
\thispagestyle{empty}
\pagestyle{empty}

\begin{abstract}
We present an active embodiment identification method for legged robots that jointly learns information-seeking behavior and explicit embodiment prediction.
Using a history-augmented URMA architecture, the method infers joint-level and global embodiment parameters through interaction with the environment in simulation across different morphologies.
\end{abstract}

\begin{figure}[!htbp]
\centering
\includegraphics[width=\linewidth]{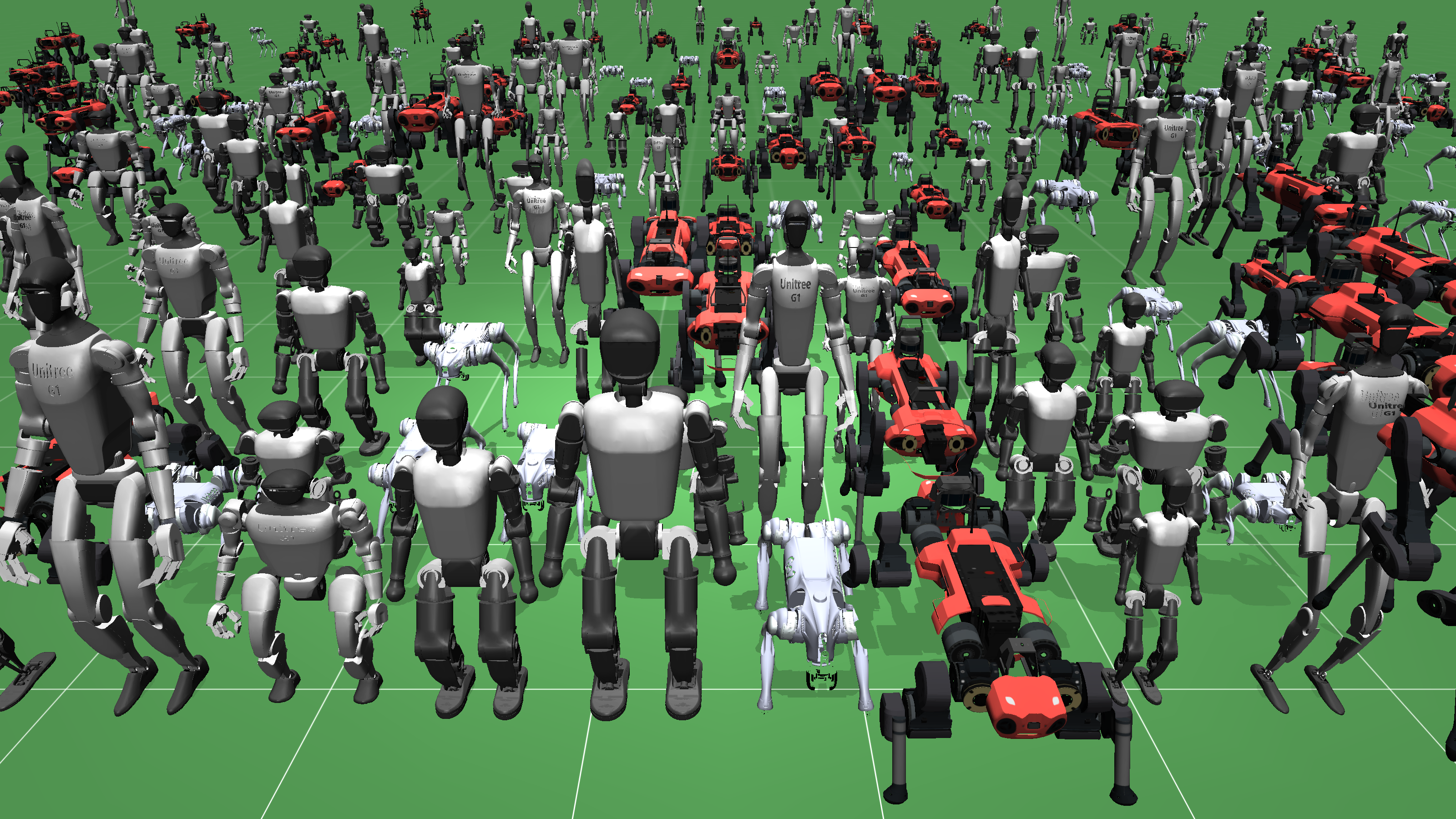}
\caption{Examples of randomized variants of the Unitree Go2 and ANYmal C quadrupeds, as well as the Unitree G1 and Booster T1 humanoid robots.}
\label{fig:hero}
\end{figure}

\section{INTRODUCTION}
Training powerful control policies for general-purpose robots is currently mostly done with either \gls{rl} trained in simulation or \gls{il} from human demonstrations.
In the latter case, there is an inherent embodiment gap between the human demonstrator and the robot, which means the policy might not learn the best possible behavior for the robot \cite{nehaniv2002correspondence, bahl2022human}.
In the former case, policies are trained with a wide distribution of domain randomization, which includes the robot's embodiment, to enable robust sim-to-real transfer \cite{tobin2017domain, ji2022concurrent}.
As policies are typically trained in an embodiment-agnostic way, the training results in a policy that is robust to different embodiments, but also conservative in its behavior and does not fully utilize the robot's maximum capabilities.
Embodiment-aware policy training can be used to close the embodiment gap in simulation by conditioning the policy on the robot's embodiment and training on a wide distribution of embodiments \cite{bohlinger2024onepolicy, ai2025towards, bohlinger2025multi}.
However, during real-world deployment, the robot's true embodiment is often not precisely known.
This can be due to the transparency and tolerances of the manufacturing process, wear and tear during operation, or even active modifications to the morphology \cite{nygaard2021real}.

In this work, we propose a method for active embodiment identification, where the robot learns to identify its own embodiment through interaction with the environment.
Compared to classic system identification methods, our method does not try solve an optimization problem to find the best fitting parameters for a predefined model \cite{ayusawa2014identifiability,grandia2018contact,schulze2025floating}, but instead learns an embodiment identification network that infers the robot's embodiment from past interactions, and is trained end-to-end with a control policy.
Compared to passive embodiment identification, where either a separate network is trained on the side to estimate the embodiment parameters explicitly \cite{li2026online}, or the policy is trained to implicitly infer the embodiment \cite{liu2025locoformer}, our method trains the policy to actively help the explicit embodiment identification.

\section{ACTIVE EMBODIMENT IDENTIFICATION}
\begin{figure*}[!htbp]
\centering
\includegraphics[width=\textwidth]{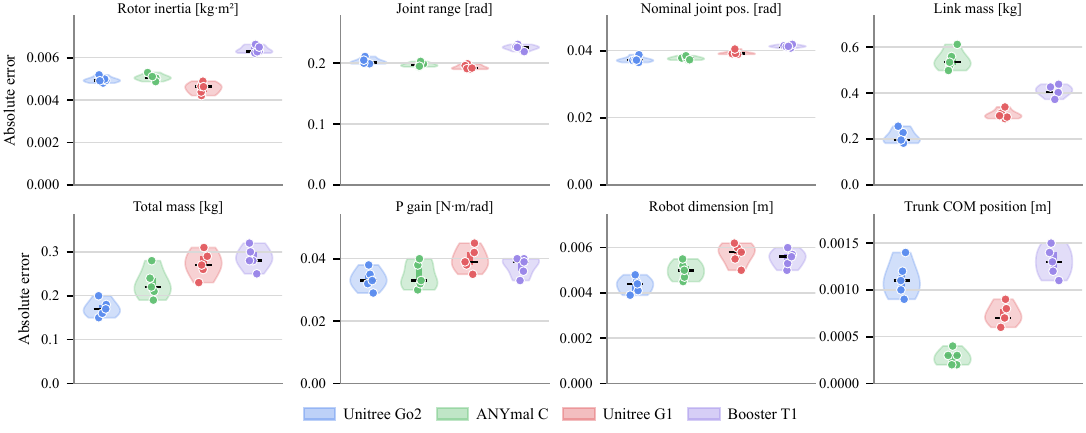}
\caption{Embodiment identification errors for joint-level (top row) and general embodiment parameters (bottom row) across four legged robots.}
\label{fig:results}
\vspace{-0.5em}
\end{figure*}

We frame embodiment identification as an active perception problem.
Unlike standard robot learning tasks, our policy's sole objective is to find information-seeking movements that excite the robot's dynamics in ways that help the embodiment identification network to accurately estimate the true embodiment parameters.

\subsection{Network Architecture}
To handle the varying action and observation spaces of different robots, we utilize and modify the \gls{urma} \cite{bohlinger2024onepolicy, bohlinger2025multi}.
Our framework consists of three networks:
\begin{itemize}
\item \textbf{Embodiment identification network:}
Uses \gls{urma}'s joint encoder to process varying numbers of joint observations, in combination with a \gls{gru} core and two separate heads to predict both joint-level (e.g., rotor inertia) and global embodiment parameters (e.g., total mass) from the interaction history.
The joint-level prediction uses an \gls{urma} decoder to produces as many joint-wise predictions as the robot has joints, while the global prediction head is a simple \gls{mlp} for a fixed-length output.

\item \textbf{Policy \& Critic:}
\gls{urma} networks that map nominal embodiment descriptions and observations to actions and values.
Both aggregate joint-wise observations into a fixed-size latent, allowing a shared architecture across robots with different numbers of joints.
\end{itemize}

\subsection{Identification Reward}
At each timestep, we calculate the \gls{mse} between the identification network's estimates and the true embodiment parameters, resulting in a joint-specific error $\mathcal{L}_{joint}$ and a general error $\mathcal{L}_{general}$. To map these unbounded errors into a normalized reward signal, we use an exponential function parameterized by a temperature $\tau$:

$$r_{id} = \frac{1}{2} \left[ \exp\left(-\frac{\mathcal{L}_{joint}}{\tau}\right) + \exp\left(-\frac{\mathcal{L}_{general}}{\tau}\right) \right]$$

By maximizing $r_{id}$, the policy learns to interact with the environment in ways that generate the most informative sequence of observations, actively driving down the identification network's estimation error.

\section{EXPERIMENTAL RESULTS}
We train the policy, critic and identification networks end-to-end in simulation for a variety of legged robots.
We include the Unitree Go2 and ANYmal C quadrupeds, as well as the Unitree G1 and Booster T1 humanoid robots.
We use \gls{ppo} \cite{schulman2017proximal} as the \gls{rl} algorithm implemented in the RL-X framework \cite{bohlinger2023rlx}, and leverage the MJX physics engine \cite{todorov2012} for simulation.
We train the policy for 2 billion steps to maximize $r_{id}$, while the robots in the 4096 parallel environments are randomized with the extreme embodiment randomization from \cite{bohlinger2025multi}.
Some examples of randomized robots can be seen in \autoref{fig:hero}.
The embodiment identification network is trained to predict the \gls{urma} joint-level descriptions, which include:
relative joint position and axis, link mass and inertia, relative COM position, number of child joints, nominal position, max torque and velocity, damping, rotor inertia, stiction and stiffness, as well as the joints' range of motion.
It also predicts the fixed-length general descriptions, which include:
control gains, action scaling factor, total mass, robot dimensions, and the mass, inertia and COM position of the trunk.

\autoref{fig:results} shows the prediction errors of the identification network for the different robots, on a select set of embodiment parameters.
The first row shows the joint-level predictions, where it can be seen that the network is able to predict the rotor inertia with an error $\approx 0.005$ kg m$^2$ for all robots, which is compared to the randomization ranges of $0.0$ to $0.015$ kg m$^2$ not very precise.
The predictions of joint ranges ($\approx 0.2$ rad error) and nominal joint positions ($\approx 0.039$ rad error) are on average more accurate, as the ranges of the joint and their positions can be between $-3.14$ and $3.14$ rad.
The identification of the general parameters seems to be significantly easier, with the total mass being predicted with an error $\approx 0.25$ kg, which is very precise when taking into account that the total mass of the robots can be up to $100$ kg for the heaviest ANYmal C variant.
Similarly, the prediction of the robots dimension and trunk COM position is even in the millimeter range.

The resulting behavior of the trained policy maintains a stable standing pose with small and medium movements in all joints.
While this prevents the robot from falling over and terminating the episode early, it also hinders the identification of parameters that define the maximum capabilities of the motors, such as the max torque and velocity, which we found to be the hardest parameters to identify.

\section{CONCLUSION}

We introduced an active embodiment identification method for legged robots that jointly learns information-seeking behavior and explicit embodiment prediction. Experiments across four robot types show that the approach can recover many embodiment parameters accurately from interaction.
This shows that active identification is a promising way to reduce the embodiment uncertainty that arises from training with domain randomization.
Promising directions for future work include feeding the embodiment predictions back into the closed-loop control \cite{li2026online}, as well as fine-tuning the embodiment-aware policies on real-world hardware \cite{bohlinger2025gait}.

\addtolength{\textheight}{-15cm}   






\bibliographystyle{IEEEtran}
\bibliography{IEEEabrv,bibliography}

\begin{thebibliography}{10}
\providecommand{\url}[1]{#1}
\csname url@samestyle\endcsname
\providecommand{\newblock}{\relax}
\providecommand{\bibinfo}[2]{#2}
\providecommand{\BIBentrySTDinterwordspacing}{\spaceskip=0pt\relax}
\providecommand{\BIBentryALTinterwordstretchfactor}{4}
\providecommand{\BIBentryALTinterwordspacing}{\spaceskip=\fontdimen2\font plus
\BIBentryALTinterwordstretchfactor\fontdimen3\font minus \fontdimen4\font\relax}
\providecommand{\BIBforeignlanguage}[2]{{%
\expandafter\ifx\csname l@#1\endcsname\relax
\typeout{** WARNING: IEEEtran.bst: No hyphenation pattern has been}%
\typeout{** loaded for the language `#1'. Using the pattern for}%
\typeout{** the default language instead.}%
\else
\language=\csname l@#1\endcsname
\fi
#2}}
\providecommand{\BIBdecl}{\relax}
\BIBdecl

\bibitem{nehaniv2002correspondence}
C.~L. Nehaniv, K.~Dautenhahn \emph{et~al.}, ``The correspondence problem,'' \emph{Imitation in animals and artifacts}, vol.~41, p.~28, 2002.

\bibitem{bahl2022human}
S.~Bahl, A.~Gupta, and D.~Pathak, ``Human-to-robot imitation in the wild,'' in \emph{Robotics: Science and Systems}.\hskip 1em plus 0.5em minus 0.4em\relax RSS Foundation, 2022.

\bibitem{tobin2017domain}
J.~Tobin, R.~Fong, A.~Ray, J.~Schneider, W.~Zaremba, and P.~Abbeel, ``Domain randomization for transferring deep neural networks from simulation to the real world,'' in \emph{International conference on intelligent robots and systems}, 2017.

\bibitem{ji2022concurrent}
G.~Ji, J.~Mun, H.~Kim, and J.~Hwangbo, ``Concurrent training of a control policy and a state estimator for dynamic and robust legged locomotion,'' \emph{Robotics and automation letters}, vol.~7, no.~2, pp. 4630--4637, 2022.

\bibitem{bohlinger2024onepolicy}
N.~Bohlinger, G.~Czechmanowski, M.~Krupka, P.~Kicki, K.~Walas, J.~Peters, and D.~Tateo, ``One policy to run them all: an end-to-end learning approach to multi-embodiment locomotion,'' \emph{Conference on Robot Learning}, 2024.

\bibitem{ai2025towards}
B.~Ai, L.~Dai, N.~Bohlinger, D.~Li, T.~Mu, Z.~Wu, K.~Fay, H.~I. Christensen, J.~Peters, and H.~Su, ``Towards embodiment scaling laws in robot locomotion,'' \emph{Conference on Robot Learning (CoRL)}, 2025.

\bibitem{bohlinger2025multi}
N.~Bohlinger and J.~Peters, ``Multi-embodiment locomotion at scale with extreme embodiment randomization,'' \emph{arXiv preprint arXiv:2509.02815}, 2025.

\bibitem{nygaard2021real}
T.~F. Nygaard, C.~P. Martin, J.~Torresen, K.~Glette, and D.~Howard, ``Real-world embodied ai through a morphologically adaptive quadruped robot,'' \emph{Nature Machine Intelligence}, vol.~3, no.~5, pp. 410--419, 2021.

\bibitem{ayusawa2014identifiability}
K.~Ayusawa, G.~Venture, and Y.~Nakamura, ``Identifiability and identification of inertial parameters using the underactuated base-link dynamics for legged multibody systems,'' \emph{The International Journal of Robotics Research}, vol.~33, no.~3, pp. 446--468, 2014.

\bibitem{grandia2018contact}
R.~Grandia, D.~Pardo, and J.~Buchli, ``Contact invariant model learning for legged robot locomotion,'' \emph{IEEE Robotics and Automation Letters}, vol.~3, no.~3, pp. 2291--2298, 2018.

\bibitem{schulze2025floating}
L.~Schulze, J.~D. Negri, V.~Barasuol, V.~S. Medeiros, M.~Becker, J.~Peters, and O.~Arenz, ``Floating-base deep lagrangian networks,'' \emph{arXiv preprint arXiv:2510.17270}, 2025.

\bibitem{li2026online}
D.~Li, B.~Ai, N.~Bohlinger, J.~Peters, H.~I. Christensen, and H.~Su, ``Online embodiment adaptation for quadrupedal locomotion,'' 2026.

\bibitem{liu2025locoformer}
M.~Liu, D.~Pathak, and A.~Agarwal, ``Locoformer: Generalist locomotion via long-context adaptation,'' in \emph{Conference on Robot Learning}.\hskip 1em plus 0.5em minus 0.4em\relax PMLR, 2025, pp. 532--546.

\bibitem{schulman2017proximal}
J.~Schulman, F.~Wolski, P.~Dhariwal, A.~Radford, and O.~Klimov, ``Proximal policy optimization algorithms,'' \emph{arXiv preprint arXiv:1707.06347}, 2017.

\bibitem{bohlinger2023rlx}
N.~Bohlinger and K.~Dorer, ``Rl-x: A deep reinforcement learning library (not only) for robocup,'' in \emph{Robot World Cup}.\hskip 1em plus 0.5em minus 0.4em\relax Springer, 2023, pp. 228--239.

\bibitem{todorov2012}
E.~Todorov, T.~Erez, and Y.~Tassa, ``Mujoco: A physics engine for model-based control,'' in \emph{2012 IEEE/RSJ international conference on intelligent robots and systems}.\hskip 1em plus 0.5em minus 0.4em\relax IEEE, 2012, pp. 5026--5033.

\bibitem{bohlinger2025gait}
N.~Bohlinger, J.~Kinzel, D.~Palenicek, L.~Antczak, and J.~Peters, ``Gait in eight: Efficient on-robot learning for omnidirectional quadruped locomotion,'' \emph{International Conference on Intelligent Robots and Systems}, 2025.

\end{thebibliography}

\end{document}